\documentclass[acmsmall]{acmart}
\renewcommand\footnotetextcopyrightpermission[1]{}
\usepackage{comment}
\usepackage{amsmath}

\usepackage{algorithm}
\usepackage{algpseudocode}

\usepackage{subcaption}
\usepackage{graphicx}

\usepackage{multirow}

\usepackage{threeparttable}
\usepackage{booktabs}
\usepackage{siunitx}

\usepackage{pgf-pie} 

\usepackage{color,array}
\usepackage{multirow}
\usepackage{svg}
\usepackage[table,xcdraw]{}
\usepackage{tabularx}
\usepackage{hyperref}
\usepackage{flushend}

\newcolumntype{C}[1]{>{\centering\let\newline\\\arraybackslash\hspace{0pt}}m{#1}}

\AtBeginDocument{%
  }

\setcopyright{none}
\begin{document}

\title{Natural Language Processing of Privacy Policies: A Survey}

\author{Andrick Adhikari}
\email{andrick.adhikari@du.edu}
\affiliation{%
  \institution{University of Denver}
  \city{Denver}
  \state{Colorado}
  \country{USA}
}

\author{Sanchari Das}
\email{sdas35@gmu.edu  }

\affiliation{%
  \institution{George Mason University}
  \city{Fairfax}
  \state{Virginia}
  \country{USA}
}

\author{Rinku Dewri}
\email{rinku.dewri@du.edu}

\affiliation{%
  \institution{University of Denver}
  \city{Denver}
  \state{Colorado}
  \country{USA}
}

\begin{abstract}
Natural Language Processing (NLP) is an essential subset of artificial intelligence. It has become effective in several domains, such as healthcare, finance, and media, to identify perceptions, opinions, and misuse, among others. Privacy is no exception, and initiatives have been taken to address the challenges of usable privacy notifications to users with the help of NLP. To this aid, we conduct a literature review by analyzing 109 papers at the intersection of NLP and privacy policies. First, we provide a brief introduction to privacy policies and discuss various facets of associated problems, which necessitate the application of NLP to elevate the current state of privacy notices and disclosures to users. Subsequently, we a) provide an overview of the implementation and effectiveness of NLP approaches for better privacy policy communication; b) identify the methodologies that can be further enhanced to provide robust privacy policies; and c) identify the gaps in the current state-of-the-art research. Our systematic analysis reveals that several research papers focus on annotating and classifying privacy texts for analysis but need to adequately dwell on other aspects of NLP applications, such as summarization. More specifically, ample research opportunities exist in this domain, covering aspects such as corpus generation, summarization vectors, contextualized word embedding, identification of privacy-relevant statement categories, fine-grained classification, and domain-specific model tuning. 

\end{abstract}



\maketitle
\thispagestyle{empty}
\pagestyle{empty}
\section{Introduction}\label{sec:intro}

Privacy policies describe an organization's data collection, use, management, and disclosure practices. 
Notably, a privacy policy should describe what personally identifiable information (PII) is collected, stored, and/or shared by the first-party and third-party organizations~\cite{zimmeck2012information}. Thus, privacy policies are vital documents that use natural languages to help users understand data access, and manage their privacy while using the services. Nevertheless, current privacy policies see limited engagement from users due to their complicated and ambiguous nature~\cite{jensen2004privacy,meiselwitz2013readability,fabian2017large,ermakova2015readability}. Such hindrances to user understanding are aggravated further by the fact that policies are often drafted with flexibility. Another challenge associated with reading policies is the requirement of significant time commitment from users~\cite{mcdonald2008cost}.

Privacy policies are also periodically revised to reflect updates in practice, making users' previous efforts on reading and understanding futile~\cite{SchwartzNC}. Additionally, the timing of a policy notification is often inopportune and leads to immediate dismissal of any notice regarding policy updates or warnings to users on data sharing~\cite{inglesant2010true}. Despite these issues, privacy policies are critical for transparency between users and service providers. These policies are also regulatory, thus mandatory for organizations to add, especially after the onset of regulations such as GDPR (General Data Protection Regulation)~\cite{voigt2017eu}, CCPA (California Consumer Privacy Act)~\cite{pardau2018california}, COPPA (Children's Online Privacy Protection Rule)~\cite{gadbaw2016legislative}, and others.

Recent advances in natural language processing (NLP) have motivated the development of applications to make privacy policies more usable. The field of NLP encompasses a variety of techniques involving computational processing to understand human languages, and can analyze documents directly, requiring minimal manual scrutiny of policy content~\cite{sadeh2013usable}. Several prior works have been published that tackle different usability aspects of policies with diverse NLP techniques. 

We have curated a collection of $109$ area relevant papers to drive research solutions toward progress on significant challenges. We provide a viewpoint of the established area through systematization and evaluation of existing knowledge and the current research landscape. Our review captures areas that have enjoyed much research attention, points out areas with unsolved challenges and presents a prioritization that can guide researchers to progress in solving fundamental challenges. 
In our analysis, we first discuss the current state of privacy policies, identify their inherent problems, and then present a brief discussion on established non-NLP research. We also highlight the gaps in non-NLP solutions that necessitates NLP application. After that, we categorize current NLP research on privacy policies into different areas: information retrieval, summarization, automatic question-answering, classification, and alignment. Finally, we study the scope of each area, identify challenges and shortcomings, and provide future research directions that can benefit from the community's attention. To our knowledge, no other survey articulates NLP research on privacy policies.  

\section{Data Collection \& Curation Process}\label{sec:methodology}
This section describes the inclusion/exclusion criteria for works explored during this review and the methodology for organization and study. 

\subsection{Material Collection}
The application of NLP on privacy policies has recently become popular. To better understand this popularity, we collected papers that study the challenges and issues of privacy policies and NLP solutions to improve the usability of privacy policies. We also curated papers that tackle the usability of privacy policies without employing NLP to stress the importance of NLP in the domain. In addition, we included papers describing developments in word embedding and NLP models, which is crucial for NLP performance on privacy policies to be at par with the current state-of-the-art in other domains. Finally, papers from domains other than privacy policies (e.g., NLP for legal documents) are also included to inspire research for usable privacy policies. 

We began by using Google Scholar as our search engine. We searched using terms such as classification, readability, usable, alignment, choice detection, change detection, vagueness, completeness, summarization, compliance, format, and others, paired with the phrase `privacy policy.' This process gave us an initial set of papers published in computer science and non-computer science conferences and journals. While computer science conferences such as USENIX, CCS, CHI, IEEE S\&P, SOUPS, WPES, AAAI, WWW, and PETS published relevant work, privacy policy-related works in computer science journals were rare. The papers were downloaded from digital libraries such as ACM Digital Library, Springer Link, and IEEE Xplore. We further expanded the list of papers by iteratively examining the references in each paper of our initial and subsequent collection of papers. Our inclusion criteria required that papers target privacy policies and their usability, and focus on improving them through non-technical guidelines or computational methods, including NLP. We collected a set of $176$ papers using this process and removed duplicates through manual scrutiny of the curated list. The final set contained $109$ papers, spanning two decades of research.

\begin{table}[]
\caption{Number of reviewed privacy policy works per topic }
\centering
\begin{tabular}{C{0.25\linewidth}C{0.45\linewidth}C{0.15\linewidth}}
\hline
\textbf{Topic} &\textbf{ Finer Topic Categorization} & \textbf{Number of Papers} \\ \hline
\multirow{3}{*}{\parbox{2cm}{\centering Comprehension challenges}} & Readability & 7 \\
 & Ambiguity & 3 \\
 & Accessibility & 4 \\ \hline
\multirow{3}{*}{Non-NLP solution} & Requirements and regulations &  5 \\ 
    & Policy design & 10 \\
 & Machine readable format & 10 \\ \hline
Data creation and analysis & - & 9 \\ \hline
\multirow{5}{*}{NLP solution} & Information retrieval & 15 \\
 & Summarization & 1 \\
 & Question-answering & 4 \\
 & Classification & 22 \\
 & Alignment & 3 \\ \hline
Word embedding model & - & 4\\ \hline
\end{tabular}
\label{tab:paperDis}
\end{table}

\subsection{Material Categorization}
We gathered a total of $109$ papers, of which $82$ were explicitly tailored to the field of privacy policies. The remaining $27$ papers were devoted to the use of NLP in general or to its use in fields similar to privacy, and are analyzed to aid in identifying research potential for NLP in the privacy domain. During our analysis, we identified five significant categories of privacy policy research: `Comprehension challenges,' `Non-NLP solutions,' `Dataset creation and analysis,' `NLP solutions,' and `Word embedding model.' These categories are further divided into sub-categories as they appear in the manuscript. Table~\ref{tab:paperDis} shows the number of papers in each of the categories mentioned earlier in the research. It should be noted that a single paper may explore multiple study areas and will belong to multiple categories.

\begin{table*}[]
\caption{Reviewed works focusing on privacy policy comprehension challenges}
\centering
\begin{tabular}{p{0.32\linewidth}p{0.1\linewidth}C{0.12\linewidth}C{0.12\linewidth}C{0.12\linewidth}}
\hline  
\textbf{Papers}                                                                & \textbf{Year} & \textbf{Readability} & \textbf{Ambiguity} & \textbf{Accessibility} \\ \hline
\citeauthor{jensen2004privacy}\cite{jensen2004privacy}             & 2004 & \checkmark          & -         & \checkmark             \\ 
\citeauthor{milne2006longitudinal}\cite{milne2006longitudinal}     & 2006 & \checkmark           & -         & -             \\
\citeauthor{mcdonald2008cost}\cite{mcdonald2008cost}               & 2008 & \checkmark          & -         & -             \\ 
\citeauthor{SchwartzNC}\cite{SchwartzNC}                           & 2009 & -           & -         & \checkmark             \\
\citeauthor{inglesant2010true}\cite{inglesant2010true}             & 2010 & -           & -         & \checkmark            \\ 
\citeauthor{meiselwitz2013readability}\cite{meiselwitz2013readability}   & 2013 & \checkmark & - & - \\ 
\citeauthor{ermakova2015readability}\cite{ermakova2015readability} & 2015 & \checkmark           & -         & -             \\ 
\citeauthor{reidenberg2015disagreeable}\cite{reidenberg2015disagreeable} & 2015 & - & \checkmark & - \\ 
\citeauthor{reidenberg2016automated}\cite{reidenberg2016automated} & 2016 & -           & \checkmark         & -             \\
\citeauthor{fabian2017large}\cite{fabian2017large}                 & 2017 & \checkmark           & -         & -             \\ 
\citeauthor{libert2018automated}\cite{libert2018automated}         & 2018 & \checkmark           & \checkmark         & -             \\ 
\citeauthor{habib2020s}\cite{habib2020s}                           & 2020 & -           & -         & \checkmark \\
\hline
\end{tabular}
\label{tab:challenges}
\end{table*}
Studies examining the usability problems with privacy policies are called `Comprehension challenges.' We further sub-categorize papers into three facets of challenges: `Readability,' `Ambiguity,' and `Accessibility.' These subjects highlight the distinctive features vital for efficient communication of privacy practices. Table~\ref{tab:challenges} provides a list of papers classified with categories of challenges. 

\begin{table*}[]
\caption{Reviewed works presenting non-NLP solutions to address privacy policy comprehension challenges}
\centering

\begin{tabular}{p{0.40\linewidth}p{0.09\linewidth}p{0.13\linewidth}p{0.12\linewidth}p{0.12\linewidth}}
\hline
\textbf{Papers} & \textbf{Year} & \textbf{Recommen-dations \& regulations}&\textbf{ Policy design} & \textbf{Machine readable policy} \\ \hline
\citeauthor{bohrer2000customer}\cite{bohrer2000customer}     & 2000 & -         & -         & \checkmark \\
\citeauthor{world2002platform}\cite{world2002platform}       & 2002 & -         & -         & \checkmark \\
\citeauthor{cranor2002p3p}\cite{cranor2002p3p}               & 2002 & -         & -         & \checkmark \\
\citeauthor{ashley2002p3p}\cite{ashley2002p3p}               & 2002 & -         & -         & \checkmark \\
\citeauthor{cranor2003p3p}\cite{cranor2003p3p}               & 2003 & -         & -         & \checkmark \\
\citeauthor{agrawal2003xpath}\cite{agrawal2003xpath}         & 2003 & -         & -         & \checkmark \\
\citeauthor{ashley2003enterprise}\cite{ashley2003enterprise} & 2003 & -         & -         & \checkmark \\
\citeauthor{Art29}\cite{Art29}                               & 2004 & -         & \checkmark & -         \\
\hyperlink{cite.multilayer}{CIPL}\cite{multilayer}                     & 2007 & -         & \checkmark & -         \\
\citeauthor{gomez2009knowprivacy}\cite{gomez2009knowprivacy} & 2009 & -         & \checkmark & -         \\
\citeauthor{kelley2009nutrition}\cite{kelley2009nutrition}   & 2009 & -         & \checkmark & -         \\
\citeauthor{pinnick2011privacy}\cite{pinnick2011privacy}     & 2011 & -         & \checkmark & -         \\
\hyperlink{cite.federal2012protecting}{FTC}\cite{federal2012protecting}     & 2012 & \checkmark                       & -             & -                        \\
\citeauthor{van2012happens}\cite{van2012happens}             & 2012 & -         & \checkmark & -         \\
\hyperlink{cite.NTA2013}{NTIA}\cite{NTA2013}                           & 2013 & \checkmark & -         & -         \\
\citeauthor{WP292014}\cite{WP292014}                         & 2014 & \checkmark & -         & -         \\
\citeauthor{azraoui2014ppl}\cite{azraoui2014ppl}             & 2014 & -         & -         & \checkmark \\
\citeauthor{iyilade2014p2u}\cite{iyilade2014p2u}             & 2014 & -         & -         & \checkmark \\
\citeauthor{schaub2015design}\cite{schaub2015design}         & 2015 & \checkmark & -         & -         \\
\citeauthor{gluck2016short}\cite{gluck2016short}             & 2016 & -         & \checkmark & -         \\
\citeauthor{strahilevitz2016privacy}\cite{strahilevitz2016privacy} & 2016 & -                               & \checkmark     & -                        \\
\citeauthor{voigt2017eu}\cite{voigt2017eu}                   & 2017 & \checkmark & -         & -         \\
\citeauthor{gerl2018lpl}\cite{gerl2018lpl}                   & 2018 & -         & -         & \checkmark \\
\citeauthor{harkous2018polisis}\cite{harkous2018polisis}     & 2018 & -         & \checkmark & -         \\
\hyperlink{cite.privacyBird}{CyLab}\cite{privacyBird}                   & 2019 & -         & \checkmark & -         \\ \hline
\end{tabular}%

\label{tab:nonNLPpapers}
\end{table*}

While some solutions concentrate on effective designs and recommendations to tailor natural language privacy policies into a more usable version, a significant portion of research has focused on developing usable notice and choice modality to present users with adequate privacy-specific information. Therefore, we categorize such solutions as `Non-NLP solutions.' These papers present ideas to tackle challenges at the core of a policy disclosure methodology. 
We further categorize `Non-NLP solution' papers into `Requirements and regulations, `Policy design,' and `Machine-readable format,' each representing a distinct method of realizing a solution. Table~\ref{tab:nonNLPpapers} lists the papers that present non-NLP solutions to enhance user privacy management. 

The third category of papers is `Dataset creation and analysis,' which focuses on corpus creation and analysis to facilitate the development of natural language processing tools for privacy policies. 

\begin{table*}
\caption{Reviewed works presenting NLP solutions to privacy policy comprehension challenges}
\centering
\begin{tabular}{p{0.28\linewidth}p{0.05\linewidth}C{0.1\linewidth}C{0.1\linewidth}C{0.1\linewidth}C{0.1\linewidth}C{0.1\linewidth}}
\hline
\textbf{Paper}                                                                    & \textbf{Year} & \textbf{Information retrieval} & \textbf{Summarization} & \textbf{Question-answering} & \textbf{Classification} & \textbf{Alignment} \\ \hline
\citeauthor{stamey2009automatically}\cite{stamey2009automatically} & 2009 & \checkmark & - & - & - & - \\
\citeauthor{galgani2012combining}\cite{galgani2012combining}       & 2012 & \checkmark & - & - & - & - \\
\citeauthor{ammar2012automatic}\cite{ammar2012automatic}           & 2012 & - & - & - & \checkmark & - \\
\citeauthor{costante2012machine}\cite{costante2012machine}         & 2012 & - & - & - & \checkmark & - \\
\citeauthor{hoffman2013semantic}\cite{hoffman2013semantic}         & 2013 & \checkmark & - & - & - & - \\
\citeauthor{zimmeck2014privee}\cite{zimmeck2014privee}             & 2014 & - & - & - & \checkmark & - \\
\citeauthor{ramanath2014unsupervised}\cite{ramanath2014unsupervised}   & 2014 & -                     & -             & -                  & -              & \checkmark         \\
\citeauthor{liu2014step}\cite{liu2014step}                         & 2014 & - & - & - & - & \checkmark \\
\citeauthor{bhatia2015towards}\cite{bhatia2015towards}             & 2015 & \checkmark & - & - & - & - \\
\citeauthor{wilson2016creation}\cite{wilson2016creation}           & 2016 & - & - & - & \checkmark & - \\
\citeauthor{hosseini2016lexical}\cite{hosseini2016lexical}         & 2016 & \checkmark & - & - & - & - \\
\citeauthor{bhatia2016mining}\cite{bhatia2016mining}               & 2016 & \checkmark & - & - & - & - \\
\citeauthor{sathyendra2016automatic}\cite{sathyendra2016automatic} & 2016 & - & - & - & \checkmark & - \\
\citeauthor{liu2016analyzing}\cite{liu2016analyzing}               & 2016 & - & - & - & \checkmark & - \\
\citeauthor{sathyendra2017helping}\cite{sathyendra2017helping}     & 2017 & - & - & \checkmark & - & - \\
\citeauthor{sathyendra2017identifying}\cite{sathyendra2017identifying} & 2017 & -                     & -             & -                  & \checkmark              & -         \\
\citeauthor{nisal2017increasing}\cite{nisal2017increasing}         & 2017 & - & - & - & \checkmark & - \\
\citeauthor{harkous2018polisis}\cite{harkous2018polisis}           & 2018 & - & - & \checkmark & \checkmark & - \\
\citeauthor{zaeem2018privacycheck}\cite{zaeem2018privacycheck}     & 2018 & - & \checkmark & - & - & - \\
\citeauthor{bhatia2018semantic}\cite{bhatia2018semantic}           & 2018 & \checkmark & - & - & - & - \\
\citeauthor{liu2018towards}\cite{liu2018towards}                   & 2018 & - & - & - & \checkmark & - \\
\citeauthor{andow2019policylint}\cite{andow2019policylint}         & 2019 & \checkmark & - & - & - & - \\
\citeauthor{yang2019xlnet}\cite{yang2019xlnet}                     & 2019 & - & - & - & \checkmark & - \\
\citeauthor{zimmeck2019maps}\cite{zimmeck2019maps}                     & 2019 & - & - & - & \checkmark & - \\
\citeauthor{sarne2019unsupervised}\cite{sarne2019unsupervised}                     & 2019 & - & - & - & \checkmark & - \\
\citeauthor{story2019natural}\cite{story2019natural}               & 2019 & - & - & - & \checkmark & - \\
\citeauthor{mousavi2020establishing}\cite{mousavi2020establishing} & 2020 & - & - & - & \checkmark & - \\
\citeauthor{torre2020ai}\cite{torre2020ai}                         & 2020 & - & - & - & \checkmark & - \\
\citeauthor{mustapha2020privacy}\cite{mustapha2020privacy}                         & 2020 & - & - & - & \checkmark & - \\
\citeauthor{bannihatti2020finding}\cite{bannihatti2020finding}                         & 2020 & \checkmark & - & - & \checkmark & - \\
\citeauthor{hosseini2021analyzing}\cite{hosseini2021analyzing}         & 2021 & \checkmark & - & - & - & - \\
\citeauthor{srinath2021privaseer}\cite{srinath2021privaseer}         & 2021 & - & - & - & \checkmark & - \\
\citeauthor{adhikari2021towards}\cite{adhikari2021towards}         & 2021 & \checkmark & - & - & \checkmark & \checkmark \\
\citeauthor{bui2021consistency}\cite{bui2021consistency}         & 2021 & \checkmark & - & - & - & - \\
\citeauthor{cui2023poligraph}\cite{cui2023poligraph}         & 2023 & \checkmark & - & - & \checkmark & - \\
\citeauthor{shvartzshnaider2023beyond}\cite{shvartzshnaider2023beyond}         & 2023 & \checkmark & - & - & - & - \\
\citeauthor{tang2023policygpt}\cite{tang2023policygpt}         & 2023 & - & - & - & \checkmark & - \\
\citeauthor{goknil2024llm}\cite{goknil2024llm}         & 2024 & - & - & \checkmark & - & - \\
\citeauthor{rodriguez2024allm}\cite{rodriguez2024allm}         & 2024 & - & - & \checkmark & - & - \\
\citeauthor{adhikari2025policypulse}\cite{adhikari2025policypulse}         & 2025 & \checkmark & - & - & - & - \\
\hline
\end{tabular}

\label{tab:nlpSol}
\end{table*}

Papers that propose NLP research in the privacy domain are categorized as `NLP solutions,' We further sub-categorize these into specific areas of NLP, namely `Information retrieval,' `Summarization,' `Question-answering,' `Classification,' and `Alignment.' 
Table~\ref{tab:nlpSol} lists the papers dedicated to each area of NLP solutions for privacy policies. It is evident from the table that the majority of the research focuses on the classification of privacy-inclined texts. Information retrieval follows classification and is crucial to present relevant information to users. 
Lastly, word embedding models must effectively capture the semantics of the policy jargon since privacy policies use specialized terminology to articulate information. 

\subsection{Review Organization}
The remainder of this review is organized in the following manner. We begin in Section~\ref{sec:readability} with a general overview of the content expected in a privacy policy document, followed by a discussion on the challenges that this expectation induces on a policy document's comprehensibility. We continue this discussion in Section~\ref{sec:nlpNeed}, introducing initial attempts (mostly non-NLP methods) to address some of the challenges, and set the motivation for using NLP in the domain. Since NLP performance depends on the quality of a language corpora, we dedicate Section~\ref{sec:corpora} to discuss the privacy policy-specific corpora that originated from the academic community. Section~\ref{sec:nlpArea} presents the body of works incorporating NLP solutions to analyze privacy policies and is divided into five subsections per the categories listed in Table~\ref{tab:nlpSol}. We also discuss the current state of word embedding models in NLP and their usage in privacy policy analysis. Following this review, Section~\ref{sec:challenges} discusses the challenges we identified, and lays out select future research directions based on lessons learned in the reviewed works. Finally, we conclude the paper in Section~\ref{sec:conclusion}.

\section{Comprehension Challenges}\label{sec:readability}
In this section, we describe the generic structure of a privacy policy in modern-day web platforms. This is followed by a discussion of the current state of usability issues in such policies.

\subsection{Coverage of a Privacy Policy}\label{sec:idealpolicy}

Websites, mobile apps, and other web products and services rely on natural language privacy policies to notify users about their privacy practices. These policies are unrestricted in terms of a well-defined structure and content. Despite that, policies are expected to cover content from the following categories~\cite{wilson2016creation}.
 \begin{itemize}
     \item\textit{First party collection/use}: how and why a service provider collects user information
     \item \textit{Third party sharing/collection}: how user information may be shared with or collected by third parties
     \item \textit{User choice/control}: choices and control options available to users
     \item\textit{User access, edit, \& deletion}: if and how users can access, edit, or delete their information
     \item \textit{Data retention}: how long is user information stored
     \item\textit{Data security}: how user information is protected
     \item\textit{Policy change}: if and how users will be informed about changes to the privacy policy
     \item\textit{Do not track}: if and how do not track signals for online tracking and advertising are honored
     \item \textit{International \& specific audiences}: practices that pertain only to a specific group of users (e.g., children, residents of the European Union, or Californians)
     \item \textit{Other}: additional privacy-related information not covered by the above categories
\end{itemize}
An ideal privacy policy should adequately reflect the following $11$ principles stated by Fair Information Practices (FIPs)~\cite{ISOPDF} for efficient privacy management.
\begin{itemize}
     \item\textit{Consent and Choice:} User consent should be taken to collect and process their data. Users should be lucidly informed of their rights and choices. A policy should explain the implications of granting or withholding consent and provide mechanisms for the users to exercise their choice.
     \item\textit{Purpose legitimacy and specification:} An organization should comply with the purpose for data collection, and their privacy policy should communicate that purpose to users with a sufficient explanation.
     \item \textit{Collection limitation}: The collection of user data should be within the bounds of the applicable law and the necessity of the stated purpose.
     \item \textit{Data minimization}: Contact with user information should be minimized and a ``need-to-know'' principle should be followed. User information should be deleted periodically.
     \item \textit{Use, retention, and disclosure limitation:} User information should be used only for the intended purpose and retained only as long as necessary. When the specific purpose expires, the information should be locked if needed to be retained.
     \item \textit{Accuracy and quality:} User information should be collected and processed accurately. Information should be verified, and the reliability of the information should be ensured.
     \item \textit{Openness, transparency, and notice}: Organizations should clearly and sufficiently communicate the policies, practices, and procedures governing user information. The communication should include the purpose of information collection, information disclosure and sharing principles, and retention and disposal practices. User choices and mechanisms to exercise them should be communicated. Users should be notified of any changes.
     \item \textit{Individual participation and access}: Users should be able to access, review, edit and delete their information in a simple, fast and efficient manner.
     \item \textit{Accountability}: All the privacy-related policies, procedures, and practices should be documented. Third-party accountability should be ensured. A privacy officer should be assigned to enforce accountability.
     \item \textit{Information security}: Organizations should protect user information's confidentiality, integrity, and availability. Information security should be guaranteed, and compliance with legal requirements and security standards should be ensured. Periodic security risk assessment and a cost/benefit analysis should be conducted. Actions and fail-safes should be implemented for any potential event.
     \item \textit{Privacy compliance}: Organizations should ensure their compliance with privacy principles. Periodic privacy audits and internal compliance checks should be conducted. In addition, a privacy risk assessment process should be developed and maintained.
 \end{itemize}

\subsection{Challenges}\label{sec:issues}
Privacy policies pose several challenges that hinder the general public from effectively utilizing policies to make informed privacy-related decisions. We identified three categories of these challenges: readability and comprehensibility, ambiguity, and accessibility~\cite{adhikari2023evolution, wagner2023evolution}.

\subsubsection*{Readability and comprehensibility}
Privacy policies are arduously long, averaging over 2500 words \cite{mcdonald2008cost}, complicated, and have low readability and comprehension. This discourages users from attempting to read and understand them. In terms of time investment, a user will have to spend at least 181 hours per year to read applicable policies~\cite{mcdonald2008cost, libert2018automated}. 

The incomprehension of policies also affects service providers. $65\%$ of online consumers decide not to register at a website because they believe that the privacy policy is incomprehensible and service providers lose valuable customers as a result~\cite{westin2004craft}. Evaluation of privacy policies with empirical readability scoring such as SMOG, RIX, LIX, GFI, FKG, ARI, and FRES metrics reveal that majority of the population cannot understand the language used in privacy policies, which requires at least a college-level reading ability \cite{jensen2004privacy,meiselwitz2013readability,fabian2017large,ermakova2015readability}. To put things in perspective, the average reading level of an adult in the U.S. is a $7^{th}$ grade reading level. 
Regulations are in place to make it easier to understand policies. Nevertheless, with the requirement to comply with multiple established regulatory requirements, privacy policies have only grown in length and complexity~\cite{milne2006longitudinal}.

\subsubsection*{Ambiguity}
The number of different ways a policy can be interpreted makes privacy policies ambiguous. In order to give organizations more flexibility, policymakers frequently use ambiguity or vagueness, thereby concealing potentially harmful practices. Even law and policy experts find it challenging to agree on decisive policy composition~\cite{reidenberg2015disagreeable, reidenberg2016automated}. The incompleteness of information also contributes to ambiguity. For example, a typical policy discloses only $15\%$ of third-party data flow practices, and $7\%$ of policies do not mention the `Do Not Track' signal~\cite{libert2018automated}. 

\subsubsection*{Accessibility}
Accessing specific portions of a policy is also challenging for many users. An average user needs help finding information regarding a specific data practice due to the tremendous effort required to read the policy in the first place~\cite{habib2020s}.
Finding policies on a website can be challenging for an average adult as well~\cite{jensen2004privacy}. In addition, the timing of privacy notices and security warnings is often inopportune, leading to immediate dismissal, as policies are shown at times that conflict with the user's primary task~\cite{inglesant2010true}.

\section{Need for Natural Language Processing}\label{sec:nlpNeed}
There are approaches other than NLP that can be used to address the challenges discussed in Section~\ref{sec:issues} and achieve usable privacy policies. However, these methods fell short and could not address the underlying issues with policies. This section discusses such approaches and their drawbacks, which prompted the research community to investigate NLP. Then, in Section~\ref{sec:nlpNeed}, we go over the various subjects of NLP research in depth.

\subsubsection*{Requirements and recommendations}

To increase policy usability, several organizations have issued recommendations and mandated requirements. For instance, the European Article 29 Working Party provides suggestions for IoT devices~\cite{WP292014}, and the National Telecommunications and Information Administration offers policy drafting instructions for mobile apps~\cite{NTA2013}. In addition, the Federal Trade Commission (FTC) in the United States suggests having privacy rules that are concise and easy to understand~\cite{federal2012protecting}. On the other hand, the General Data Protection Regulation (GDPR) mandates greater transparency in privacy policies for data processing in Europe~\cite{voigt2017eu}.

The effectiveness of a policy is determined by when it is provided, how it is delivered, the type of interaction used, and how to control the choices given. \citeauthor{schaub2015design} \cite{schaub2015design} have identified requirements and best practices for practical and usable privacy notice design. They provide an overview of a design space to consider while authoring comprehensive policies that meet the audience-specific requirements and incorporate the constraints of policy practices. This design space contains dimensions such as the timing of notices, dissemination channels, the modality of communication, and actionable controls.

\subsubsection*{Policy redesign}
To improve the accessibility of conventional privacy policies, numerous extra or modified formats of expressing policy practices are suggested, given that natural language policies are required and cannot be replaced.
\begin{itemize}

    \item \textit{Short notices}: To effectively communicate vital information to consumers without overwhelming them, policies can be condensed~\cite{gluck2016short, strahilevitz2016privacy}.  
    
    \item \textit{Multi-layered policies}: The Art.29 Data Protection Working Party suggested multi-layered privacy policies that place the most important information first (directly visible to all)---the controller's identity, the aims of processing, and circumstantial procedures. The second and third layers would provide interested parties with more thorough information~\cite{Art29}. The Center for Information Policy Leadership at the law firm of Hunton \& Williams made one of the initial attempts to build multilayer privacy notification standards~\cite{multilayer}.
  
    \item \textit{Graphical privacy policies}: Graphical representations of policies for increasing user understanding are also proposed. Privacy icons can represent data types and practices, such as $\mathsf{KnowPrivacy}$  icons~\cite{gomez2009knowprivacy}. To better convey the level of privacy protection and invasion, icons can be supplemented with colored rings~\cite{pinnick2011privacy}. Similar colored symbol indicators are used by browser plugins like $\mathsf{Privacy Bird}$ to show whether a website's P3P policy (a machine-readable format)  complies with the preferences of the user~\cite{privacyBird}.
    The tabular privacy format known as `privacy nutrition labels' is another graphical representation that can help users better understand privacy policies by making the information more readily available, and by tabulating information on data collection, usage, and sharing~\cite{kelley2009nutrition}.
\end{itemize}

\subsubsection*{Machine-readable policies}
In order to be automatically processed by computers, machine-readable policies require XML or another computer-readable language. The World Wide Web Consortium's specification of P3P (Platform for Privacy Preferences) was the most reviewed attempt~\cite{cranor2003p3p}. Statements from the P3P XML specification describe the procedures for handling data sets. Each statement includes the data categories, the intended use, the recipients, and the retention policy. However, due to their intricate definitions, the privacy taxonomy and language in the XML specification proved controversial~\cite{world2002platform}. There are also a variety of P3P extensions available, but they received little traction~\cite{cranor2002p3p, ashley2002p3p, agrawal2003xpath}.
    
Following the introduction of P3P, new languages with comparable syntax were developed, including the ``Enterprise Policy Authorization Language'' (EPAL) \cite{ashley2003enterprise}, ``Accountability Policy Language'' (A-PPL) \cite{azraoui2014ppl}, ``Customer Profile Exchange'' (CPExchange) \cite{bohrer2000customer},  ``Purpose-to-Use'' (P2U) \cite{iyilade2014p2u} and ``Layered Privacy Language'' (LPL) \cite{gerl2018lpl}. These languages make some improvements in terms of usability and enforcement while being similar to P3P.

\textit{Requirements and recommendation} in reality make privacy policies more complicated. For example, the GLBA and HIPAA Privacy Rules governing finance and healthcare, respectively, the COPPA rule governing children's information, the Video Privacy Protection Act (VPPA), and several other U.S. national security laws, all impose constraints on how privacy policies are written. Additionally, data imported into U.S. organizations must comply with a ``Safe Harbor'' agreement or foreign regulation. As a result of trying to simultaneously meet all the standards and serve the interests of the organization, regulators, and customers, these regulations result in extremely complicated privacy policies.

\textit{Policy redesign} solutions suffer from vagueness and incompleteness. Alternatively, the general public will only be informed partially if privacy warnings are condensed only to include the most crucial procedures. Furthermore, multi-layered policies allow businesses to choose each segment's language and content, which may conflict with user expectations of critical information. For lay people, graphical visualizations like nutrition labels and privacy icons are devised, yet they show ambiguity in interpretation. The same is valid for privacy icons, where ambiguity might be introduced by cultural, educational, or contextual variations.

\textit{Machine-readable policies} lack a workable and scalable implementation, making it challenging to employ them in actual applications. The lack of acceptance and human comprehensibility is another problem for machine-readable formats. Furthermore, many Internet users need help utilizing the P3P software that comes pre-installed in many systems. Another issue is that neither users of the Internet nor websites are required to utilize machine-readable formats. As a result, machine-readable privacy policies are no longer considered as helpful or practical as once.

Many of the shortcomings of one format can be remedied by another. For instance, privacy icons or nutritional labels may be used to make natural language privacy policies easier to read. To exemplify, \citeauthor{harkous2018polisis} merged natural language and graphical privacy policies, improving the legal value of a natural language privacy policy with the accessibility of policy icons~\cite{harkous2018polisis}. Similarly, \citeauthor{van2012happens} employed a tiered strategy to blend graphical and natural language policies~\cite{van2012happens}.

All of the aforementioned issues regarding usable privacy policies are amplified due to the frequent changes a policy undergoes. Each change will require a user to re-evaluate a policy in its entirety, with or without an accompanying usability enhancement tool. In general, succinct yet precise summarization of changes in a privacy policy is yet to see much research interest.

In summary, natural language privacy policies are a popular way to ``notice and choice.'' However, many early works focus on alternative formats to communicate privacy practices and hence suffer from poor adoption or lack of expressiveness. Methods such as privacy nutrition labels have received attention from a select few large organizations (e.g., Google and Apple). However, natural language documents remain their primary, detailed, privacy-related communication tool. This has directed recent research to embrace natural language processing as the method of choice to extract relevant privacy information from a policy document, design query systems, and summarizations, and identify presentation issues that hamper a user's ability to comprehend such documents. NLP solutions do not require organizations to reformat their mode of privacy communication and work directly on documents that already exist in most organizations.

\section{Privacy Policy Corpora}\label{sec:corpora}
The development of machine learning and natural language processing methods depends on the availability of corpora of domain-specific texts. Like in other domains, an NLP model for privacy policies has to learn the syntax and semantics of the language typically used in such policies to identify privacy-specific artifacts effectively. Supervised approaches that further require the learning objective to be exemplified in the corpora demand that manual annotations on the text are also available. Quality privacy policy corpora are critical in this research domain.  

\subsubsection*{OPP-115} \citeauthor{wilson2016creation}  generated a dataset called the  OPP-115 corpus that contains policy information from 115 websites~\cite{wilson2016creation}. With annotation for 23,000 data practice statements, 128,000 practice attributes, and 103,000 annotated text spans, OPP-115 is the most extensively used corpus. The OPP-115 corpus policies are annotated with the ten high-level categories listed in Section \ref{sec:idealpolicy}.
High-level categories are further broken down with a set of attributes that are distinct from each other. For example, a `User choice/control' data practice is associated with attributes such as `choice type,' `choice scope,' `personal information type,' `purpose,' and `user type.'
`First party collection/use' and `Third party sharing/collection' are the two most frequently occurring categories in this corpus, hinting at the focus of most policies. The `Other' category also gets heavily used, potentially indicating disagreements between annotators resulting from the ambiguity in privacy documents. Categories such as `User choice/control,' `User access, edit, and deletion,' and `Do not track' appear with a relatively lower frequency, indicating a paucity of information for user-specific actionable items.

\subsubsection*{PPCRAWL} Over the past twenty years, there have been more than twice as many privacy policies. Because past studies had only been able to examine privacy policies from a single moment in time, \citeauthor{amos2021privacy} created a longitudinal collection of policy corpora using a crawler on Archive's Wayback Machine~\cite{amos2021privacy}. This collection now contains 1,071,488 English language privacy policies. Policies from 130,000 websites, some of which have been around for more than 20 years, are included in this corpus. Broken links and careful processing to obtain the relevant texts are challenges when building a corpus with automated crawlers~\cite{wilson2016creation,amos2021privacy, sarne2019unsupervised,bannihatti2020finding}.

\subsubsection*{PRIVASEER} The PrivaSeer corpus is a large, single snapshot, English language corpus of 1,005,380 privacy policies from 995,475 different web domains, gathered around early August of 2019~\cite{srinath2021privaseer}. The corpus is created using a crawler seeded with 3.9 million potential URLs to privacy policies; the downloaded content was further filtered using language detection, document classification, URL cross-verification, and duplicate removal. The PrivaSeer corpus is also indexed and can be searched through a web interface.

\subsubsection*{OPT-OUT-236} Another crucial component of rules that provide consumers control over data collection and use is the ability to opt out. However, it is challenging to locate these options in a policy. Thus, \citeauthor{bannihatti2020finding} created a corpus of 236 website privacy policies to study the automatic extraction of opt-out statements from privacy policy text~\cite{bannihatti2020finding}. 3,213 hyperlinks were extracted and labeled using the Document Object Model (DOM) of policies, utilizing the top 500 U.S. Alexa websites, Selenium~\cite{Selenium}, Geckodriver~\cite{Geckodriver}, and Mercury Parser API~\cite{Mercury}. The annotations have 441 links to third-party service opt-outs and 2,692 links to first-party opt-out choices.

\subsubsection*{APP-350} The APP-350 corpus contains policies of Google Play Store apps that have more than 50 million installs and randomly chosen apps with more than 5 million installs, totaling 350 policies. The corpus was created to study enhancements that can be made to mobile application policies~\cite{zimmeck2019maps}. 

\subsubsection*{PRIVACYQA} PRIVACYQA is another corpus of mobile application policies created for evaluating automated question-answering systems~\cite{ravichander2019question}. This corpus has 3,500 annotations of appropriate responses to 1,750 questions identified by experts.

Numerous other corpora are created expressly for study or advancement in specific NLP fields. For instance, a corpus of 400 policies annotated with risk levels for an email address, credit card number, social security number, advertisement and marketing, location, personally identifiable information of children, sharing with law enforcement, policy change, control of data, and data aggregation have been used in research for automated risk assessment~\cite{zaeem2018privacycheck}. Another example of one of these corpora is the 130,326 Android application policies collection, which was utilized to create a privacy domain-specific word-embedding~\cite{harkous2018polisis}.

A perusal of current corpora reveals that generating annotated privacy policy corpora is time-consuming and frequently calls for the assistance of topic specialists. However, crowdsourcing can be a potential workaround to obtain high-quality annotations that address scalability issues. Furthermore, when annotating texts as ambiguous as in privacy policies, inter-annotator disagreements can further make it difficult to establish ground truths~\cite{wilson2016creation, zimmeck2019maps, ravichander2019question}.

\section{Natural Language Processing of Privacy Policies} \label{sec:nlpArea}
By engineering computational models and methods, natural language processing can be applied to human language problems, such as in privacy policies. Topics, including pertinent information extraction, content summarization, automated question-answering, document categorization, and document clustering, are all covered in the NLP application field for privacy policies. In reality, the themes mentioned above overlap, but we discuss the works within the context of a single field at a time. Towards the end of this section, we also briefly review the word embedding models in general and expand upon the models that have been tried and tested for privacy policies.

\subsection{Information Retrieval}
Information retrieval systems are designed to assist users in finding crucial data in the most practical format at the precise moment they need it~\cite{kenter2017neural}. Information retrieval from privacy policies entails identifying the texts' sections that meet the needs of the current task. If done correctly, information extraction from policies can reduce the work required of users to comprehend the subtleties controlling their privacy.

Several challenges appear when designing a system to capture the syntactic and semantic information encoded in a text document. Syntax in privacy policies encompasses words and their arrangement in a sentence, conforming to formal grammar rules. Semantics is about the universally coded meaning, and pragmatics encoded in words and how an audience interprets them.

\subsubsection*{Semantic information extraction}
Any written sentence uses a specific word arrangement, known as semantics, to convey some meaning. Formal semantics, lexical semantics, and conceptual semantics are just a few branches and sub-branches of semantics study. The logical components of meaning, such as reference, implication, and sense, are described by formal semantics. Word relations are described by lexical semantics, and cognitive structure is described by conceptual semantics. 

A defined, finite set of terminology that incorporates interconnected semantic concepts, and is utilized in knowledge management, is called an ontology. An information type ontology construction technique was demonstrated by~\citeauthor{hosseini2016lexical} using a manual grounded analysis of five privacy rules~\cite{hosseini2016lexical}. This approach was tested against 50 mobile privacy policies, resulting in an ontology comprising 355 distinct pieces of information. The method is directed by seven heuristics that are used to extract associations between hypernyms, meronyms, and synonyms from information-type phrases, and finally culminated into 14 semantic rules. The semantic rules can be expanded to improve extraction efficiency; however, such an approach needs to adapt to the dynamics of policy composition and needs a clear strategy for information extraction that occurs automatically. 

Ontology generation can assist in contradiction detection in privacy statements by capturing both positive and negative data collection statements within privacy policies. $\mathsf{PolicyLint}$ is one such tool, which uses an expanded set of Hearst patterns~\cite{hearst1992automatic} on named-entity recognition, parts-of-speech analysis, and type dependence, to extract ontologies for both data objects and entities~\cite{andow2019policylint}. To extract a succinct representation of the grammatical links between the data objects, entities, and verbs, $\mathsf{PolicyLint}$ builds data and entity dependency (DED) trees using dependency parse trees. When building a DED tree, paths between labeled nodes are calculated on a dependency-based parse tree by copying nodes associated with a manually curated lists of verbs applied in sharing and collection statements, data objects, and entities, while retaining information about negated verbs and exception clauses. Contradictions and limited definitions are detected using predicate logic rules on the resulting four tuples actor, action, data object, and entity components.

When $\mathsf{PolicyLint}$ was used to analyze 11,430 privacy policies from well-known apps on Google Play, it was found that $14.2\%$ of the statements contained logical conflicts and $17.7\%$ contained narrowing definitions. A closer look found false presentations and redefining of popular terminology, which are alarming outcomes. 

Another computational linguistic method used to analyze the connections between documents and terms is latent semantic analysis (LSA), which generates a set of related concepts. The method relies on a size-optimized word count matrix generated from the documents, the optimization being primarily obtained using singular value decomposition (SVD), which can then be analyzed using vector similarity measures to estimate document similarity. By automatically identifying the most important subjects of a privacy policy and, as a result, the most critical words of those topics, LSA helps highlight the underlying semantic links between words in privacy policies~\cite{stamey2009automatically}. 

Inferred semantic relations can also shed light on ambiguity and characterize semantic ambiguity. For example, words with a limited number of contexts are less semantically ambiguous than words that appear in a wide variety of situations on various themes. LSA has also been applied to determine the level of semantic variance, also known as a word's semantic diversity (SemD)~\cite{hoffman2013semantic}. Values were higher for words that appeared in various contexts and vice versa. As an alternative, the $\mathsf{Hermes}$ prototype system by~\citeauthor{stamey2009automatically}  identifies ambiguities based on semantic similarity measures between a user's policy and signatures derived from a typical privacy policy \cite{stamey2009automatically}. 

Semantic connections can also highlight the shortcomings of privacy policies. For instance, by representing data practice descriptions as semantic frames, one can analyze the degree of integration of semantic roles with data action within a frame~\cite{fillmore1976frame}. ~\citeauthor{bhatia2018semantic} used this approach on 202 annotated statements from five privacy policies, yielding 17 semantic roles and 281 instances of data actions~\cite{bhatia2018semantic}. They determined whether the data practice description was incomplete by looking at missing role associations. It was noted that almost $32\%$ of statements about retention, $45\%$ of statements about sharing, and $19\%$ of statements about usage lack topic roles and purpose roles. 

While specialized approaches exist that use syntax-driven semantic analysis methods to construct partial ontologies, and context-free grammar for inferring semantic relations~\cite{hosseini2021analyzing}, deep learning and NLP can facilitate automated methods for improved and scalable extraction of semantic frame representations of policies, and enable large-scale analysis. Shvartzshnaider et al. proposed information extraction through semantic role labeling (SRL) using domain-specific rule-based heuristics to include information for a predefined list of verb predicates~\cite{shvartzshnaider2023beyond}. Adhikari et al. demonstrated the value of such fine grain information extraction in its ability to serve as building blocks for a variety of other tasks, including the creation of alternative visualizations and question-answering systems~\cite{adhikari2025policypulse}. Among other applications of SRL in the privacy domain, \textsf{PurPliance}~\cite{bui2021consistency} has utilized it to handle lengthy and complex phrases within purpose clauses. However, these predicates are mostly limited to first-party collection and use of data, and rarely include information beyond the highly coupled application-specific requirements.

\subsubsection*{Entity extraction}
The main focus of entity extraction for privacy policies has been extracting lexicons and keywords that serve as placeholders for essential data in a privacy policy. In order to extract entities automatically, entity extraction often necessitates the establishment of a manual vocabulary and the use of a parts-of-speech (POS) tagger. However, because manual annotations quickly reach saturation, it is only possible to generate a complete vocabulary of lexicons~\cite{bhatia2015towards}.

Using stop words to separate the text into candidate keywords, a general unsupervised keyword extractor such as Rapid Automated Keyword Extraction (RAKE) also enables entity extraction from individual documents. The degree and frequency of the word vertices in the word co-occurrence graph are then used to grade potential entities~\cite{rose2010automatic}. Entities resembling catchphrases can also be extracted using conditional rules based on statistical information~\cite{galgani2012combining}. Named Entity Recognition (NER) is another method to identify collected data, entities collecting information, served purposes, and subsumption relations, thus aggregating the information dispersed across a policy~\cite{cui2023poligraph}. These extracted entities provide concise information, but needs detailed and accompanying textual descriptions. Entity extraction can help categorize privacy policies or extract features, but relationship links must be established between the extracted keywords to guide privacy-specific actionable recommendations. The non-availability of annotations, and coupled scalability issues, also place restrictions on entity extraction research.

Combining crowdsourcing with natural language processing can be a successful strategy for extracting entities from privacy regulations. When paired with crowd worker annotations, dependency tree parsing can be utilized to identify actions on various information kinds reliably~\cite{bhatia2016mining}. State of the art in NLP still needs to improve in human interpretation, even though it offers a practical way to scale the extraction from more documents. Crowdsourcing and NLP complement each other and fill in each other's gaps. While NLP struggles to find semantic traits for meaningful information extraction on its own, crowd workers are prone to missing information.

\subsection{Summarization}
Summarization can condense privacy policies to critical points by finding elements of interest in the document and encapsulating the most important content. Summarization can either be extractive or abstractive. Extractive summarization extracts, simplifies and organizes sentences in a document to convey critical information. Abstractive summarization conveys information in a document through abstract generation~\cite{jurafsky2000speech}. 

The only summary tool for privacy policies is $\mathsf{PrivacyCheck}$, providing an extractive summary in the format of $10$ user-essential questions with extracted answers~\cite{zaeem2018privacycheck}. The questions were compiled using surveys and expert consultations, and answers were extracted using a classification model trained using the Google Prediction API on a corpus of 400 policies. $\mathsf{PrivacyCheck}$ also provides a risk score based on the extracted answers.

\subsection{Question-Answering}
Question-answering (QA) extracts pertinent words, phrases, or sentences from a document in response to a request. For instance, if users want to inquire about data storage, they can ask, ``How long will my information be stored within the organization?'' The system should then be able to extract the policy segments that correspond to the user's query. Finding specific material areas is imperative for these systems that work with natural language documents.

There are two versions of QA, open and closed. In closed QA, questions are marked with predefined labels. Annotations in a corpus such as OPP-115 can define value level, attribute level, and category level granularity for question labels and can then be used to realize a closed QA~\cite{sathyendra2017helping}. However, testing showed that it is challenging to map every user inquiry to one of the predetermined categories.

In the area of privacy policies, where queries are not given labels, open QA methodologies have more potential. Privacy policy segments are broken into smaller segments, and the pertinent parts are recovered using similarity scoring on segment embedding. With the help of query expansion and length reduction techniques, the Bi-LSTM Attention-based Deep Neural Model can anticipate similarities between a user's question and potential responses to pick the best answer~\cite{sathyendra2017helping}. Predicate logic for structured querying can also be created using segment classification output~\cite{harkous2018polisis}. 

Finding question-answer pairs using a tree-edit distance-based model is an alternative strategy. In this method, text pairs are first transformed into dependency trees, and the degree of similarity is determined by the number of edits necessary to convert the tree for a question into the tree of an answer~\cite{heilman2010tree}. SVM ranking can also be used to determine similarity in answer retrieval~\cite{kim2015convolutional}. This was adopted to develop a legal question-answering system for the Japan Civil Code~\cite{do2017legal}. However, the use of these methods in privacy policy analysis is yet to be seen.

The interaction with large language models (LLMs) is inherently in a question-answer format. As such, their use in extracting information from a (often long) privacy policy in this naturally appealing manner has started to gain traction. The design space of such approaches are centered around prompt engineering, where comparative advantages of specifying scoped policy text, its placement in a prompt, splitting of prompts, output format instruction, templating, chain-of-thought prompting and few-shot training, are evaluated~\cite{goknil2024llm,rodriguez2024allm}. With careful design and fine-tuning, LLMs have been shown to be able to perform at par with statistical approaches in identifying specific types of personal data collection, and often better than traditional symbolic approaches in identifying contradictions. Advancing LLM performance in policy analysis tasks can address scalability and generality barriers, which are often found in narrowly scoped symbolic and statistical methods. 

\subsection{Classification}
The availability of machine learning models and privacy policy corpora has sparked a significant amount of research into the automated classification of policies. Text classification is a traditional application of NLP and can be used for various purposes in the field of privacy policies.

\subsubsection*{Supervised segment classification} Supervised text classification is a technique used to process documents written in a natural language. This is heavily demonstrated in the legal regime, where classifiers were employed for the automatic recognition of arguments in legal writings and extraction of features involving the lexical, syntactic, semantic, and discourse qualities of texts~\cite{moens2007automatic, francesconi2007automatic}. Furthermore, introducing neural networks in the domain has significantly improved text classification performance~\cite{chen2015convolutional}.

\citeauthor{ammar2012automatic} demonstrated the viability of extracting salient characteristics from privacy policies and evaluating whether a concept is present in a policy by training a logistic regression classifier on a limited sample of privacy policies~\cite{ammar2012automatic}. The categorization of privacy regulations was eventually tested using several alternative classifier models, including k-NN, SVM, LSVM, and decision trees~\cite{costante2012machine}.

The job at hand often determines how well a classifier model performs. While some models are more effective at classifying segments (paragraphs), other models might be more effective at identifying the presence of a notion. For instance, the multinomial naive Bayes classifier was the most appropriate when determining the presence of `collection,' `encryption,' `ad-tracking,' `limited retention,' `profiling,' and `ad-disclosure' concepts in a privacy policy~\cite{zimmeck2014privee}. On the other hand, SVM outperformed models such as logistic regression and the hidden Markov model for automated classification of policy segments with categories from the OPP-115 corpus~\cite{wilson2016creation}. The embedding utilized to represent the policy text is just as essential as the model; compared to logistic regression and convolutional neural networks, TF-IDF vectorization with SVM improved segment and phrase classification using categories in the OPP-115 corpus~\cite{liu2018towards}. 

\subsubsection*{Domain-specific embedding} It is critical to note that the language used in policy writings is often specialized, making it impossible for a general word embedding model to represent it adequately. Using word embedding produced from training on policy texts is a crucial way to get around this limitation. For example, with domain-specific embedding, a policy analysis tool called $\mathsf{Polisis}$ was developed that classifies both high-level privacy practices and fine-grained data in privacy policies, using a hierarchy of convolutional neural networks~\cite{harkous2018polisis}. 

Subsequently, BERT (Bidirectional Encoder Representations from Transformers)~\cite{devlin2018bert}, a transfor-mers-based deep learning model with domain-specific word embedding, performed much better than previously benchmarked CNN-based models to classify segments~\cite{mousavi2020establishing,srinath2021privaseer}. 
Furthermore, by adopting modeling techniques from autoencoder models (like BERT) while avoiding their constraints, extended autoregressive language models such as XLNet~\cite{yang2019xlnet} can work with unsupervised representations of text sequences and can surpass baselines set with BERT~\cite{mustapha2020privacy}. The availability of pre-trained models that can be fine-tuned with the downstream task, such as training a custom word embedding, is an added benefit of using models like BERT and XLNet. As a result, these models are often considered state-of-the-art in privacy policy classification.

\subsubsection*{Sentence classification} While the availability of well-performing text classification models has made privacy policies more amenable for high-level concept extraction, note that a policy segment may contain statements from multiple categories. When training classification models, these sentences may inject noise into the semantic interpretation of a category. However, most research has been restricted to segment categorization due to a lack of sentence annotation corpora, and only segment annotations are available in gold standard corpora such as OPP-115.

Although some studies have looked into categorizing sentences, they are primarily interested in identifying the existence of a specific notion. One example is determining if a sentence implies a choice regarding described privacy practices. By converting sentences into unigram and bigram bag-of-words features and then modifying those features to indicate the presence of modal verbs and opt-out specific phrases, a logistic regression model that outperforms linear SVM, random forest, naive Bayes, and nearest-neighbor models for opt-in/out choice detection can be trained~\cite{sathyendra2016automatic}. Active learning for data set cleaning and upgrading into a two-classifier architecture improves choice detection performance, where the first classifier determines whether a sentence is a choice instance, and the second classifier automatically identifies and labels different types of opt-out choices offered in privacy policies~\cite{sathyendra2017identifying}. To improve the performance of a baseline bag-of-words model, a combination of feature types such as stemmed unigrams and bigrams, relative location in the document, and opt-out specific phrases were used. The tools created from these works were also incorporated into a browser extension~\cite{nisal2017increasing}. For creating tools for opt-out detection and making the results available to users, logistic regression is preferable to BERT~\cite{devlin2018bert} and FastText~\cite{joulin2016bag} when factors such as words and bigrams, modal verbs/key phrases, subject modeling, hyperlink URL, and hyperlink anchor text are taken into account during feature creation~\cite{bannihatti2020finding}.

Sentence classification is also used in compliance studies involving apps, focusing on categorizing sentences of policies in the ecosystem of Android applications~\cite{story2019natural, zimmeck2019maps}. \citeauthor{story2019natural} used classification and static code analysis of apps to identify discrepancies in practice by comparing the projected policy label values to the APIs used within a program. The APP-350 corpus's policies were used to train, validate, and test the classifier. The findings show a variety of non-compliance, with cookies, device IDs, and mobile carrier identifiers being the most notable examples. Potential compliance issues relating to location and third parties are also quite important.

Low-frequency categories, such as `Do not track,' have been found to perform poorly in multi-class classifications~\cite{wilson2016creation,liu2018towards}. In addition, due to a lack of context, sentence-level classification fared poorly when compared to segment-level classification in such categories. According to experimental findings, ambiguity affects how privacy policies are automatically classified since the imbalance between categories often hampers classifier performance.

\subsubsection*{Regulatory compliance} Another application of supervised classification has been demonstrated in AI-enabled completeness testing of privacy rules in compliance with the General Data Protection Regulation~\cite{torre2020ai}. At three distinct hierarchical levels, texts were classified using ML-based classifiers (SVM), similarity-based classifiers (cosine similarity), and keyword-based classifiers. Sentences were vectorized using GloVe~\cite{pennington2014glove}, and text generalization swapped out specialized textual elements for more general ones. Through the development of a conceptual model, many tiers of metadata types were created to describe the information content that GDPR intended for privacy rules. Automatic metadata is employed to determine if a given policy complies with the GDPR's information requirements.

\subsubsection*{Unsupervised classification} Supervised classification of privacy policies requires text annotations, preferably from law experts who are conversant with the language used in such documents. However, this can be highly expensive and needs to scale better. As an unsupervised alternative, the intersection of language learned from topic modeling, and category-specific essential vocabulary can also be used to connect privacy practices stated in a document to categories and themes~\cite{liu2016analyzing}. For example, \citeauthor{sarne2019unsupervised} proposed an unsupervised method for classifying policy texts by using Latent Dirichlet Allocation (LDA) on 4,982 privacy policies~\cite{sarne2019unsupervised}. LDA produced word clusters for a pre-set limit of $100$ topics, and word probabilities concerning each topic were calculated. The topic with the highest probability is given to a paragraph after the topic probabilities for each word in the paragraph are added up. Although the $100$ identified topics were found to cover only $36$ specific subjects (through manual processing), the topic model analysis suggested the existence of more detailed categories beneath each high-level category compared to the annotations in the OPP-115 corpus.

Recent work in unsupervised classification attempts to leverage the power of LLMs in text analysis, and make inferences about a text segment's privacy-specific category label. \textsf{PolicyGPT} is one such attempt where multiple LLM models were evaluated on their labeling accuracy against two human annotated corpus~\cite{tang2023policygpt}. The primary challenge here is to clearly establish the context of the task, which includes communicating the descriptions of the categories and engineering the prompts to guide the process. Models such as GPT4, even in a zero-shot configuration, demonstrated the potential to carry out the classification with reasonable accuracy; however, the performance can vary significantly across the data sets used for evaluation. Further, the efficacy of this approach in performing finer grain sentence classification is unknown. 

We note that unsupervised classification of policy text segments is often stated as a method to obtain automated annotation on a large corpus of privacy policies~\cite{goknil2024llm}. However, in the absence of a subsequent process to evaluate the correctness of the assigned labels, the procedure is still subject to the usual concerns of precision and recall, and not to be used as ground truth in downstream tasks. 

\subsection{Alignment}
The task of splitting a given text data set into topically coherent pieces is known as an alignment or text segmentation. In order to improve user accessibility and comprehension, alignment seeks to generate parts where each segment only covers one topic. For the alignment of policy segments, an unsupervised model based on a hidden Markov model (HMM) may be used. The model's parameters are virtually identical to those of a traditional HMM, except that each hidden state in the model corresponds to a topic and that emissions are represented by multinomial distributions rather than categorical ones~\cite{ramanath2014unsupervised, liu2014step}.

The use of text segmentation is not just restricted to privacy policies. In order to align segments that are semantically related, LDA was used to gather thematic data about the segments~\cite{misra2009text}. A dynamic programming technique that discards irrelevant segments can be utilized for alignment. However, LDA suffers from poor processing speeds, and privacy policy segments often contain text relating to multiple themes.

$\mathsf{GRAPHSEG}$ is another technique that can be used to align sentences that are semantically connected to one another~\cite{glavavs2016unsupervised}. A semantic relatedness graph is constructed in this method, where nodes represent sentences and edges are placed between nodes based on the cosine similarity of two sentences. Coherent segments are then identified using a maximal clique finding algorithm. This method does not assign predefined topical labels to text segments but instead attempts to identify sentences that elaborate on the same topic. 

\subsection{Word Representation Models}

Word embedding (or representation) maps text into a continuous vector space with a fixed dimension. The assignment of a similar vector representation to related words is crucial in any natural language domain. Following the distributional notion that similar words tend to appear in similar situations, most embedding models are constructed for this purpose using co-occurrence statistics from sizable monolingual corpora. However, many NLP vectorizations regard words as atomic units represented as indices in a lexicon (for instance, one hot encoding) with no concept of word similarity. Representing the semantics of a text mathematically is still a challenge; word embedding techniques that can capture semantic, syntactic, and thematic information are still the subject of ongoing research. Nevertheless, the use of NLP in the privacy sector relies heavily on word embedding.

The most typical fixed-length vector form for texts is the bag-of-words or bag-of-n-grams~\cite{harris1954distributional}. Since bag-of-words do not encode the word order, representations of distinct sentences can be identical. Furthermore, it has a high degree of dimensionality, and due to sparsity in the data, bag-of-words, and bag-of-n-grams essentially do not encode semantic information.
A fundamental framework for learning vector representation using a recurrent neural network (RNN) was outlined by ~\citeauthor{bengio2003neural}~\cite{bengio2003neural}. The RNN was trained to predict the next word using the previous words as input. This strategy dramatically outperforms n-gram models~\cite{brown1992class} and enables the use of lengthier contexts. Furthermore, training a neural network on such vectorized word representations makes it possible to realize a more straightforward model to learn continuous word vectors. There are two architectures in this regard: the continuous bag-of-words model, which predicts the current word based on context, and the skip-gram model, which predicts surrounding words based on the current word. Using these architectures, embeddings are generated on large corpora using Word2Vec~\cite{mikolov2013efficient}. Word2Vec has also been used to train privacy-specific word embeddings for extrinsic tasks such as question-answering, albeit the effectiveness of these word embeddings was not compared to generic word embeddings~\cite{sathyendra2017helping}.

Analysis of Word2Vec's model features reveals fine-grained semantic and syntactic regularities, subsequently used to propose GloVe. This new global log-bilinear regression model combines global matrix factorization with local context window techniques~\cite{pennington2014glove}. In addition, GloVe uses globally aggregated word-word co-occurrence counts, while Word2Vec is guided by statistics of words that frequently occur besides a given term.

Using an extension of the continuous skip-gram model, another method is to learn representations for character n-grams and express words as the sum of the n-gram vectors. Consequently, this results in the inclusion of subword information in the process. By merging the vectors of each word's constituent subwords, this model, known as FastText\footnote{https://github.com/facebookresearch/fastText}, can be applied to words that are absent from the corpus~\cite{bojanowski2017enriching}. FastText word embeddings trained with privacy policies outperform generic embeddings like GloVe~\cite{harkous2018polisis, kumar2019quantifying}. Furthermore, it was discovered that higher dimensional embeddings frequently yield better results than their lower dimensional counterparts. Despite requiring additional training time due to its higher order and a more significant number of dimensions, the higher order dimension may expressly capture interactions between words.

Recent advances in deep learning models using transformers-based mechanisms have produced effective models like BERT that accurately capture the contextual relationships between words and subwords~\cite{devlin2018bert, vaswani2017attention}. An encoder reads the transformers' text input, followed by a decoder that forecasts the task output. As the complete string of words is read at once, capturing both the left and right context of any word, BERT is inherently bidirectional in encapsulation contexts. However, autoregressive learning models, including BERT, are restricted to taking context into account either in a forward or a backward direction and not both at once. Permutation language modeling was the solution XLNet suggested for learning the bidirectional context~~\cite{yang2019xlnet}. All token permutations are taken into account for any sequence of tokens. As a result, the model will learn to gather data from both forward and backward directions through permutation.

Another contextualized word representation, called Embeddings from Language Models (ELMo) \cite{peters2018deep}, simulates words' syntactic and semantic aspects that depend on their context. These word vectors are learned functions of a deep bidirectional language model's internal states (biLM). Word-level embeddings are produced using a bi-directional LSTM (long short-term memory) using chara-cter-level tokens as inputs.

In policy classification for classes with a more significant number of examples, BERT performed admirably~\cite{kumar2019quantifying}. However, BERT performs worse for classes with fewer examples. This is in line with the findings of numerous NLP tasks, including question-answering and natural language inference. However, this reported behavior is from an off-the-shelf model trained on the OPP-115 corpus for three epochs, and no fine-tuning of any hyper-parameters for BERT was performed. Nevertheless, a thorough exploratory investigation of the BERT model's use in the field of privacy regulations may produce workable solutions for privacy-related applications.

\section{Challenges and Future Work}\label{sec:challenges}
NLP applications in privacy policy analysis are gaining traction in the research community. We consider these applications primarily exploratory at this point but have been instrumental in revealing challenges and possibilities. We briefly outlined some issues in specific approaches in Section~\ref{sec:nlpArea}. This section provides a summative overview of the fundamental building blocks of NLP and usable privacy policy that researchers may undertake to push beyond prototypical examples.

\subsection{Information Retrieval}
\subsubsection*{Privacy-specific parser} Prior research in information retrieval for privacy regulations has mainly relied on rule-based heuristics, which have limitations when addressing texts that do not follow a clear pattern. The majority of the effort involves extracting features from the text, such as dependency tree parsing, word occurrence statistics, and POS tags, and then applying a rule to the features to retrieve the necessary information. Because privacy-specific terminology is present, using a generic parser to analyze privacy messages results in incomplete information being captured. These methods also suffer from information retrieval saturation, which results in outcomes that are not full, as is the case with lexicons~\cite{bhatia2015towards}.

\subsubsection*{Using neural networks} Information retrieval tasks have been handled utilizing neural networks for natural language processing in different fields. Researchers have used CNNs to pinpoint occurrences, the arguments supporting those events, and the roles those arguments played in those events~\cite{chen2015event}. There is also an RNN-based model to determine the cause of such incidents~\cite{nguyen2016joint}. An event schema can also be induced for open domain events using a latent variable neural model~\cite{liu2019open}. Adopting neural network-based approaches for privacy policies may also overcome the limitations of a rule-based approach and be a possible direction for future research.

\subsubsection*{Corpus generation} Another significant challenge is creating a knowledge base, which requires intensive manual effort and is highly specific to the task at hand. Developing a primary corpus that can be used to realize multiple information retrieval tasks from a privacy policy can be a valuable asset to the research community. Large corpora such as PPCRAWL and PrivaSeer are now available but do not contain ground truth annotations. 

\subsubsection*{Change detection} Given the effort required to absorb the variety of content present in a privacy policy, and the frequent updates they undergo, automated detection of changes in privacy practices can facilitate a better-informed user. For example, methods such as dependency parse trees help identify the simple noun, pronoun, or verb changes in equivalent statements across two policies \cite{adhikari2021towards}, but the creation of a practice change summary is yet to be attempted. Preference-based filtering can also be incorporated to tune a policy's view, or the changes, to severity levels and features of interest to a user. 

\subsection{Summarization}
\subsubsection*{Summarization vectors} Although summarization of privacy policies can be beneficial to users, it is one of the most under-researched privacy policy areas. It was concerning to see that only one out of the $103$ analyzed papers discussed summarization~\cite{krantz2018abstractive}. An extractive summarization of privacy policies relies on the selected questions, similar to a question-answering application, but with a fixed set of questions. The disadvantage of extractive summarization is that the results heavily depend on the questions chosen and how they are framed. This may result in the loss of important privacy policy texts. Another issue is that the results are unaltered texts from the privacy policy itself, so the overall language may need to be clarified for a user. Alternatively, summarization that aims to extract the `who,' `what,' `why,' `when,' and `how' of privacy practices can present a concise yet user-oriented overview of a policy document.

\subsubsection*{Domain-specific summarization} An abstractive summarization can give users more useful information by presenting the policy more uniformly. The privacy policy field has yet to take a step towards abstractive summarization, and the works comparing various models for abstractive summarization~\cite{krantz2018abstractive} can serve as a good starting point for future research into an abstractive summarization tool for privacy policies. 

\subsection{Question-Answering}
\subsubsection*{Contextualized word embedding} A similarity metric between the word vectors of a query and the segments is presently used in question-answering work to retrieve pertinent policy segments. As a result, the kind of embedding model used for the job significantly impacts the application's performance. FastText-created domain-specific embedding outperformed generic embedding in terms of performance~\cite{harkous2018polisis}. However, FastText is a static word embedding method, meaning that once it has been learned, the context is not altered by the embedding, and the embedding does not change across sentences. Since user inquiries are not restricted to a fixed format and given that a word's representation might alter depending on the context in which it occurs, contextualized word embedding may lead to improved outcomes.
Thus, exploring contextualized word embedding can positively contribute to building more efficient QA systems. 

\subsubsection*{Large language models} Research in privacy policy analysis through LLMs is fresh and yet to be established. LLMs provide a question-answering format that is amenable to human interaction. Approaches to align analysis tasks to this format are taking shape, but the effectiveness of LLMs in parsing through the complex verbiage of a privacy policy document remains to be studied. As such, the interpretation of policy text as made by a LLM, and the completeness of the extracted information is still open to assessment, much like in other methods. The inference process of LLMs is also not well-understood, and can present a challenge when it comes to linking answers to sections of policy text.

\subsection{Classification}
\subsubsection*{Category standardization} Although classification is the area of NLP research on privacy policies that have received the greatest attention, an ideal model or set of features still needs to be implemented. The approaches taken for categorizing a privacy policy are wholly reliant on the work at hand, and the best option can only be found through experimentation. Implementing a multiple-categorization architecture that can be applied to any application with privacy policies can be advantageous. Additionally, finer categories are necessary to realize such a classification architecture. 

\subsubsection*{Fine-grained categories} A possible step towards a complete granular classification of a privacy policy document is to use a hierarchical classification with OPP-115 categories at the top level and finer-grained categories below.
The absence of data with annotations is a problem while developing such a system. As shown by \citeauthor{sarne2019unsupervised}~\cite{sarne2019unsupervised}, the OPP-115 corpus includes annotations that only partially cover all the privacy notions. Most of the ongoing research in the field may benefit from the thorough identification of more granular categories and the corresponding annotation, if done correctly.

\subsubsection*{Sentence classification with context} Another problem with privacy policy classification is the lack of context during prediction. Because sentence-level classification lacks the context customarily included in the paragraph-level classification, its performance is diminished. However, classifying sentences at the paragraph level runs the danger of classifying a sentence belonging to a distinct class with the majority class of the paragraph. This problem can be solved by turning a sentence into a word vector, doing sentence-level classification \cite{adhikari2022sentence}, and using contextualized word embedding to take the context of the entire paragraph into account.

\subsection{Alignment}
\subsubsection*{Performance optimizations} Even though HMM-based techniques outperformed topic models and clustering, text alignment is a computationally expensive job~\cite{ramanath2014unsupervised, misra2009text, glavavs2016unsupervised}. However, categorizing phrases before alignment can still considerably boost performance. When employing classification to carry out an initial clustering, which might reduce the number of comparisons, \citeauthor{adhikari2021towards} demonstrated improvements in sentence matching~\cite{adhikari2021towards}. Similar intra-category alignment may also enhance the performance of current alignment techniques. Using categories also allows for parallelizing prediction tasks, dividing them, and aligning intra-categorically.

\subsection{Word Representation}
\subsubsection*{Fast word embedding} Language modeling with neural networks has several benefits. However, contrary to commonly used n-gram language models, these networks use implicit smoothing. Because the full lexicon is projected onto a thin hidden layer, semantically related words cluster together. Although several studies have attempted to solve this issue, the time required to train such big models (with millions of parameters) is a recognized drawback of modern word representation models~\cite{levy2015improving}.
Without external resources, a linear post-processing transformation method can enhance the semantic and syntactic information encoded in the embedding, as demonstrated by intrinsic and extrinsic evaluation of similarity and relatedness for the changed embedding~\cite{artetxe2018uncovering}.

\subsubsection*{Domain-specific model tuning} Hyperparameter tuning is mainly responsible for the performance improvements of neural network word embedding models. These modifications can be applied to conventional models to achieve comparable performance improvements, demonstrating that one method has no overall advantage over another~\cite{levy2015improving}. The same conclusion holds for the privacy policy field; there is no right or wrong embedding model, and practical model tuning may produce the best outcomes for any given task. Instead of building features and examining their effects, future work in the privacy policy domain may explore basic models and fine-tune these models themselves.

\subsection{Overall Analysis}
Our investigation revealed that most NLP-based efforts were placed on categorizing policy contents into categories about privacy practices. Classification occasionally deals with different issues, such as ensuring compliance. Information extraction, summarization, question-answering, and alignment are all strategies that try to convey privacy-focused information to consumers in the most understandable way possible.
The main issue with the current research trend is the separation of aims. The ideal course would be to have a unified privacy framework that can enable categorization, summarization, alignment, question-answering, and information extraction on a shared basis. Automating a natural language policy into a machine-readable format is one way to achieve a common foundation supporting various NLP applications. Current research also focuses on developing technologies to make privacy policies usable across all business areas. Nevertheless, in practice, depending on the nature of the business, policies might vary considerably. For instance, privacy policies communicating practices of a social media organization are articulated differently than privacy policies referring to banking or financial domains. The performance of currently available tools is limited due to these differences in characteristics that still need to be addressed.

\section{Conclusion}\label{sec:conclusion}
Prior research has looked into how NLP can be used to create automated systems for better readability and transparency of privacy policies. We conducted a detailed systematic review of $103$ peer-reviewed academic works to organize this research. 
We found that most studies concentrate on addressing one aspect of a privacy policy, such as identifying the theme of a paragraph, and do not consider the various other factors required for a  comprehensive solution. 
Additionally, most proposed methods suffer from a lack of adoption due to high computational requirements. 
Most approaches frequently require manual analysis and evaluations to validate the outcomes, significantly slowing down the development process. 
As we realized from our review, there is ample scope in this research domain while focusing on the user side. 
Applying NLP in privacy policies generally presents several difficulties. However, due to the permanence of natural language policies, NLP is the only option for providing users with helpful information. 
Researchers must investigate how to give consumers brief notices that are dynamically customized for each person. We conclude by proposing that we need user interfaces that human analysts can utilize to accurately and consistently define a policy. Moreover, such interfaces should provide the most relevant answers to users within the bounds of acceptable computation. 
The output of NLP on privacy policies must be multidimensional and be able to address a wide range of privacy-related questions.

\bibliographystyle{ACM-Reference-Format}
\bibliography{references.bib}

\end{document}